# DEEP MULTIPLE INSTANCE LEARNING FOR FORECASTING STOCK TRENDS USING FINANCIAL NEWS

Yiqi DENG and Siu Ming YIU

Department of Computer Science, The University of Hong Kong,
Hong Kong, China

***ABSTRACT***

*A major source of information can be taken from financial news articles, which have some correlations about the fluctuation of stock trends. In this paper, we investigate the influences of financial news on the stock trends, from a multi-instance view. The intuition behind this is based on the news uncertainty of varying intervals of news occurrences and the lack of annotation in every single financial news. Under the scenario of Multiple Instance Learning (MIL) where training instances are arranged in bags, and a label is assigned for the entire bag instead of instances, we develop a flexible and adaptive multi-instance learning model and evaluate its ability in directional movement forecast of Standard & Poor's 500 index on financial news dataset. Specifically, we treat each trading day as one bag, with certain amounts of news happening on each trading day as instances in each bag. Experiment results demonstrate that our proposed multi-instance-based framework gains outstanding results in terms of the accuracy of trend prediction, compared with other state-of-art approaches and baselines.*

## 1. INTRODUCTION

Stock trend prediction has always been a hotspot for both investors and researchers to facilitate making useful investment decisions, conducting investment, and gaining profits. Normal trend prediction tasks mainly take direct views on the stock prices. Based on stock prices, fundamental analysis [1], technical analysis [2, 3], and historical price time series analysis [4-6] have been used to aid in previous stock analysis. In addition to the direct quantitative information the numeric price brings on stock trends, financial news implies qualitative relations between daily events and their effect on the stock prices. Intuitively, people intend to buy stocks on hearing positive news and sell on negative news. Literature in [7-9] has also indicated that events reported in financial news play important roles concerning the stock trends in the financial market.

Using financial news to predict stock trends can be regarded as one text binary classification task. Take one trading day as an example, the trend of the stock is up if the closing price of the day is higher than the previous day, otherwise, it will have a downward trend. However, the uncertainty in daily financial news presents challenges to our normal financial text analysis. The financial news uncertainty on each day comes from two sources: uncertainty in the news amounts happening on each day and uncertainty in the number of positive and negative news each day. For the uncertainty in daily financial news occurrence, financial news appears randomly most of the time. Sometimes there is no relevant news in one day while sometimes there can be more than ten or twenty news in one day (see Figure 1). As it can be seen from Figure 1, the number of news articles appearing each day, each month presents certain randomness. Sometimes there is no relevant news in a day, and sometimes there are hundreds of news articles in a day. In view of the uncertainty in the number of positive and negative news each day, the stock trend in a day is generally related to a certain amount of specific financial news instead of one piece. On a day with stock trends going up, it is quite unrealistic to consider that all the news within this day is positive, in that there can be lots of positive news, as well as some negative news, neutral or even unrelated news on this day.

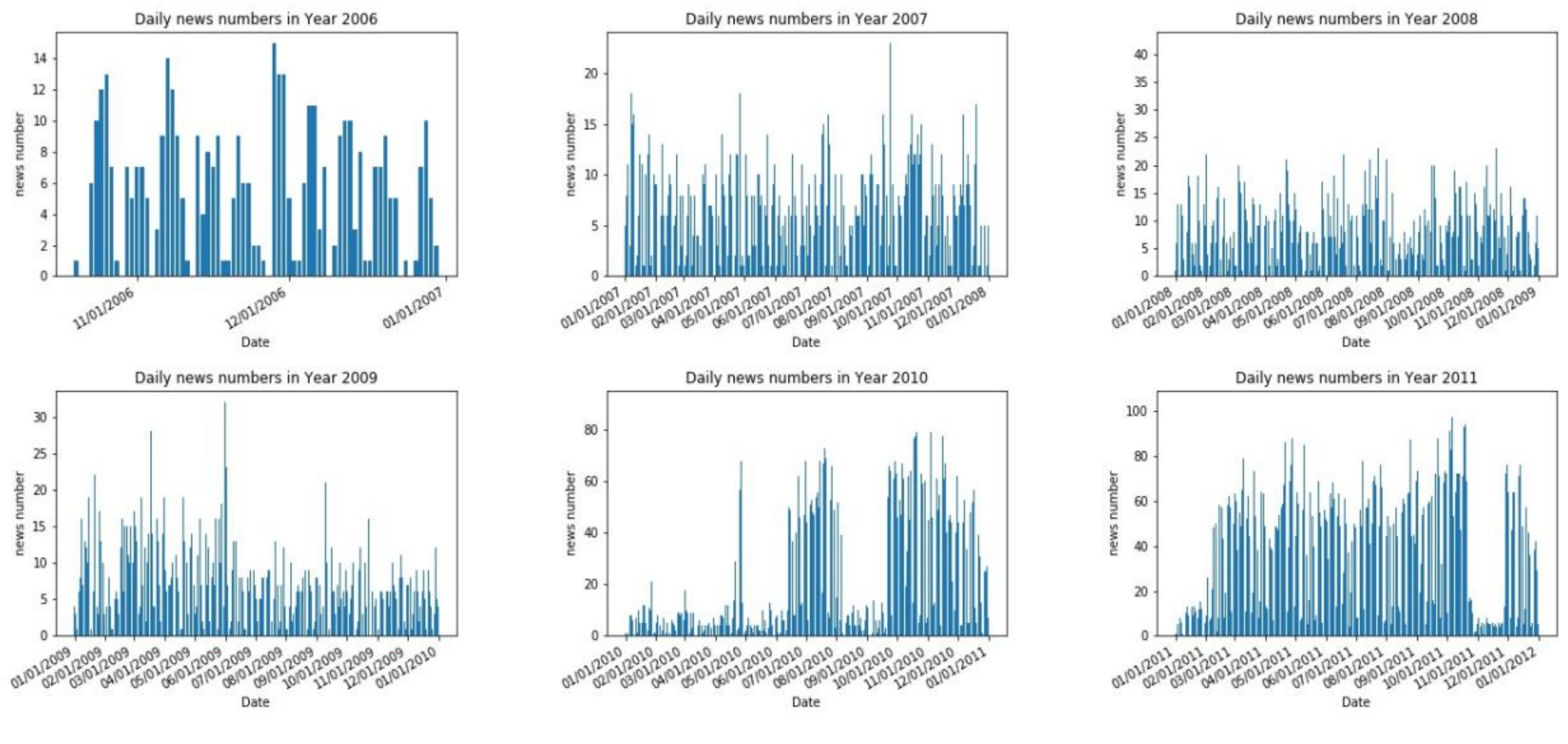

Figure 1. Random appearing news amounts in each year.

A good example can be illustrated through a direct look at the news published on a trading day (See in Table 1, Figure 2). There're almost 100 pieces of news on September 20, 2011. The stock closing price is lower than that of the previous day, which we considered a downward trend on this trading day. In all the news on this day, some are conveying a positive signal, say 'boost', 'increase', while some are sending neutral or even negative aspects ('cost', 'falter', 'low') to investors. The uncertainty on the distribution of positive and negative news within one trading day is seldom discussed in the previous studies. The reason behind this lies in the lack, intact labelings of single news. Indeed, considering the rapid changes in news amounts and random occurrences, the labeling of individual news (instance labels) is quite expensive and impractical. Besides, investors with different risk preferences treat and mark differently on the aspects of each piece of news without uniform standards, which also increases the difficulty of labeling individual news. However, the weak unknown news label relationships are indirect yet hardly negligible. Sometimes, a sudden piece of good news could alter investors' previously bearish decisions when investors do tradings. Therefore, ignoring news labels can do harm to precise stock prediction.

At the most of time, we can only gain class labels for groups of news in a day (bag-level labels) instead of each piece of news (instance-level labels). In the scenario of classifying stock trends using one day's news, labels for groups of news within a day (bag labels) are reflected as the annotation of stock trend, which can be clearly and easily clarified through the change of two consecutive stock closing prices. With full-annotated, complete daily group class labels, supervised learning has dominated in previous literature [10-15] pertaining to stock trend forecasting tasks through financial news. However, few current studies consider the randomness occurrence of financial news and include weak unknown news label relationships within a day in their modeling due to the lack of individual news (instance) labels.

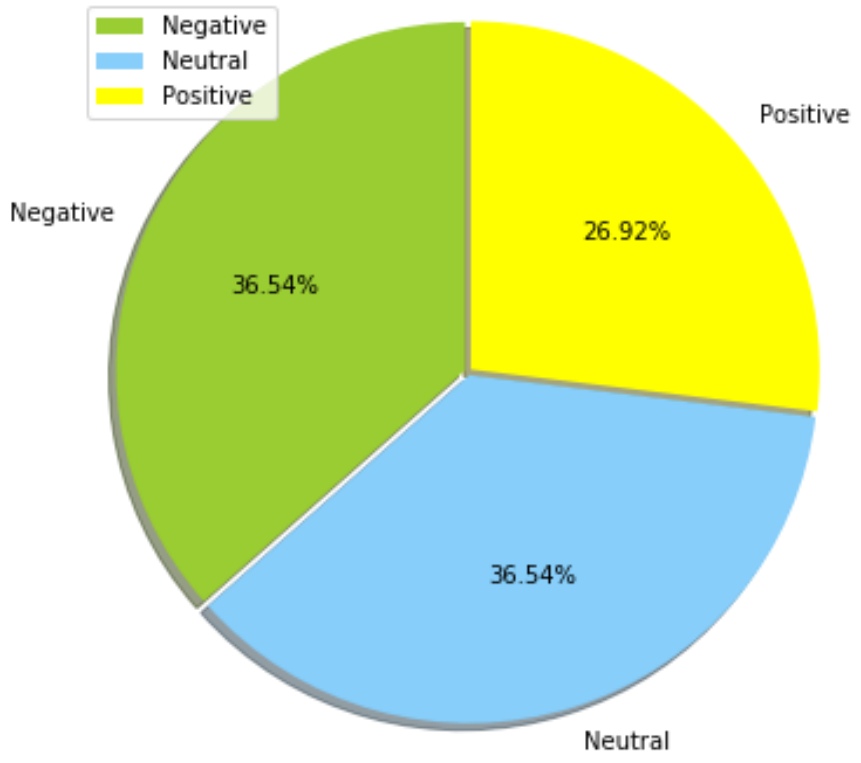

Figure 2. Sentiment values of news items on September 20, 2011, calculated by Natural Language Toolkit (NLTK)

Table 1. News posing on Sep. 20, 2011

| | **News** |
|---|---|
| ++ (pos) | China's Stocks Rise From 14-Month Low; Commodity Producers Gain. |
| ++ (pos) | Obama's Home State Illinois Turns to China for Economic Boost. |
| ++ (pos) | U.S. Gulf Crude-Oil Premiums Increase as Brent-WTI Gap Widens. |
| ++ (pos) | U.S. Natural Gas Fund Premium at 0.31% on Sept. 19 |
| … | |
| - - (neg) | China Endorsing Tobacco in Schools Adds to $10 Trillion Cost. |
| - - (neg) | China Stock-Market Sentiment at Historic Low, Citigroup Says. |
| - - (neg) | Hurricane Irene Cost NYC at Least $55M: Official. |
| - - (neg) | Oil Slides in New York on Speculation Demand to Falter; Brent Erases Drop. |
| … | |
| + - (neu) | Oil Trades Near a Three-Week Low in New York; Brent Crude Climbs in London. |
| + - (neu) | Short-Term Stimulus Won't Help U.S. in Long Run: Glenn Hubbard. |
| + - (neu) | U.S. August Building Permits by Type and Region. |
| + - (neu) | U.S. Solar Power Rises 69 Percent, Led by Commercial Projects. |
| + - (neu) | China Jan.-Aug. Average Export Prices Rise 10.3%. |
| … | |
| **Trend** | DOWN TREND |

Actually, as one type of weakly supervised learning algorithm, multiple instance learning (MIL) can be utilized to infer unknown news (instance) labels and the weak correlation between them. It helps ameliorate the limitation on the uncertainty of financial news mentioned above and models the financial news dataset at the instance, bag levels. Hence, in this work, we aim to adopt *Multiple Instance Learning* (MIL) [16] and consider the effects of financial news on stock trends from the perspective of Multiple Instance Learning. Related to the earlier work, in this paper we make the following contributions:

• A summarization of MIL principles used in scenarios of stock trend prediction using the financial news.

• Build up a novel MIL model to alleviate the problems of finance news uncertainty and insufficient individual news labels within one day when predicting the stock trends.

• Empirical evidence that our proposed MIL model can achieve impressive results on the S&P 500 stock index prediction, competing with other conventional neural architectures and previous MIL methods.

The paper is organized as follows. We review related work and earlier mainstream approaches in news stock trend prediction in Section 2. In Section 3, we introduce our proposed multiple instance learning (MIL) framework, with a description of how to represent news(instance), how to deduce the possibility of constituent news(instances), as well as the day(bag) vectors and day(bag)-level supervision. Section 4 presents our experimental results and discussion by using the financial news datasets. We also compare our method against previous approaches in this section. Finally, we conclude and summarize the paper in Section 5.

## 2. RELATED WORK

### 2.1. Multiple Instance Learning

Multiple instance learning was originally introduced by Dietterich et al., [17] in investigating the problem of drug activity prediction. In multiple instance learning, the training set comprises labeled‘bags’, each bag is a collection of unlabeled instances. The exact label of every training instance is unknown, instead, the labels are provided for the entire bag. The appearance of multiple-instance learning is gaining interest by researchers, since in the real world, on account of tedious annotation by hand, limited label sources gained, there are a variety of classification problems where class labels are not complete at the instance level but only available for groups of instances. Alleviating the burden of obtaining limited-labeled datasets, Multi-instance learning has been successfully put into practice in areas of image classification [18, 19, 20], document modelling [21], event extraction [22], sound event detection (SED) [23], etc. The multi-instance learning approach also shows its feasibility in the application of text mining tasks. He Wei et at., [24] treat each document as a bag, the sentences in the document as each instance, to investigate text classification problems. They use Bag of Sentences (BOS) as text representation. Dimitrios et al., [25] adopt multi-instance learning on the problem of predicting labels for sentences given labels for reviews. They put forward learning classifiers to predict at the instance level instead of at the bag level. Based on instance-level similarity and group-level labels, a novel objective function 'Group Instance Cost Function'(GICF) is proposed to encourage smoothness of inferred instance-level labels. Nikolaos et al., [21, 26] introduce a weighted multiple-instance regression (MIR) framework for document modeling and instance predicting aspect ratings in reviews. The MIR model captures meaningful structural information which is helpful for text understanding tasks and increases the performance of lexical and topical features for review segmentation and summarization. Stefanos et al., [27] present a neural network model for fine-grained sentiment analysis within the framework of multiple instance learning. Without the need for segment-level labels, their neural model is trained on document sentiment labels and learns to predict the sentiment of text segments. It can be indicated from the above literature that multiple instance learning, even though imperfect labels are employed, can nonetheless be used to create strong predictive models. However, in the field of financial news text analysis for stock trend prediction, little research has been done with this challenging yet potentially powerful variant of incomplete supervision learning.

## 2.2. Financial news for stock trend prediction

Financial news plays an important role with respect to the stock trends in the financial market. By means of deep learning and natural language processing (NLP), existing methods on stock market prediction by analyzing financial news have proven to be quite effective.

Financial news contains useful information in unstructured textual form. When representing each news title, it is non-trivial to extract semantic information and context information within each news title. The vector representation of words [28, 29] facilitates feature extraction from not only words but also sentences and documents. Classical methods such as averaging word vectors [30], training paragraph vectors [31] can be efficient, yet they have been indicated incapable of preserving semantics and gaining interpretation of linguistic aspects such as word order, synonyms, co-reference in the original news. To overcome this limitation, some improved representation techniques have been advanced in the following studies. Ding et al. [32] use open information extraction (Open IE) to obtain structured events representations in news. Later in [33], he put forward a novel neural tensor network to extract events in financial news. Get inspired by work [34], works such as [35, 36] adopt hierarchical structures to perform the classification: Hu et al. [35] adopt a hierarchical structure called Hybrid Attention Networks (HAN) to catch more features and help address the challenge of low-quality, chaotic online news. Liu et al. [36] extract news text features and context information through Bidirectional-LSTM. A self-attention mechanism is applied to distribute attention to most relative words, news and days. Ma et al. [37] develop a novel Distributed Representation of news (DRNews) through creating news vectors that describe both the semantic information and potential linkages among news events in an attributed news network. News vector representation has achieved state-of-art performances on various financial text classification tasks. A better text representation on news titles is vital in financial news analysis to capture features related to stock trends forecasting.

As predictive methods, deep learning models present high performances in traditional natural language processing tasks, namely, Convolution Neural Network (CNN) [38-40], Recurrent Neural Network (RNN) [41, 42], etc. In recent studies, authors in [30] propose a recurrent convolutional neural network (RCNN) model on stock price predictive tasks. Word embedding and sentence embedding are made as better embedding vectors for each piece of news. Huy et al. in [43] utilize a new Bidirectional Gated Recurrent Unit (BGRU) model for the stock price movement classification. Xu Y et al. [44] propose a stock price prediction model with the aid of news event detection and sentiment orientation analysis, through introducing Convolutional Neural Network (CNN) and Bi-directional Long Short-Term Memory (Bi-LSTM) in their predictive model. Most recently, a recurrent state transition model, integrating the influence of news events and random noises over a fundamental stock value state, is constructed in [45] for the task of news-driven stock movement prediction. A tensor-based information framework for predicting stock movements in response to new information is also introduced in [46]. From neural-network-based approaches to hierarchical structures-based models, to tensor-based networks, these methods have grown the mainstream and state-of-art techniques in the field of stock trend prediction through financial news texts.

# 3. METHODOLOGY

In this section, we describe the framework of our proposed multiple instance learning model. To further, we relate how to obtain news (instance) representations to better extract keywords and context information within the news, as well as how to apply multiple instance learning to address some of the pitfalls mentioned in previous parts, which are: the uncertainty of news relating to the randomness occurrences and the unknown annotation for each piece of news. The model design has 4 stages: word embedding, news(instance) encoding, news (instance)-level classifiers, and bag-level representation and final classification.

## 3.1. Definitions & Formulation

According to the principle of multiple instances learning (MIL), given an input dataset $D$, the dataset $D$ contains a set of labeled bags $B = \{B_1, B_2, \ldots, B_M\}$, where each bag is a collection of unlabeled instances. In our multiple instance learning (MIL) framework, we regard news as instances and all the news that appears on that day as a bag. Now we consider the prediction of the stock trend over M trading days, M trading days represent M bags in $D$ . Each bag $B_k, k = 1,2,\ldots,M$, contains $n_k$ pieces of news, where each news text(instance) $n_k^i \in R^d, i = 1,\ldots,n_k, k = 1,\ldots,M$ is a d-dimensional vector learning from neural networks. With numerical labels $\mathrm{Y_k}, \mathrm{k} = 1,\ldots,\mathrm{M}$ derived from the daily stock close price, we are given bag labels for the stock trends each day. Then we have:

$$D = \{(B_k, Y_k)\}, k = 1,2,\ldots,M$$

where $B_k \in B$ and $\mathrm{Y_k}$ is a bag label assigned to day $k$. We assume binary classification in this paper, then we have $Y_k \in \{0,1\}$, where 0 represents a downward stock trend for the day $k$, where the stock close price of current day $k$ is lower than the previous day $k-1$, and 1 shows the upward stock trend, where the stock closing price of current day $k$ is higher than the previous day $k-1$.

Previous studies mainly bring into focus on the relevance of news. They divide news into related or unrelated parts. Indeed, each piece of news conveys some of the information that drives the stock price trend, up or down. Therefore, our model attempts to predict how likely each piece of news is to move the stock upwards or downwards. The philosophy of multi-instance learning is to build classifiers to predict the labels of unknown bags by analyzing the label-known bags and its multiple instances. Based on that, in our work, we promote the relevance of news to the inference of individual news probability of being up and down.

## 3.2.Proposed Model

### 3.2.1. Word embedding

To obtain vector representation of each news text (instance), one key step is the use of embedding techniques. The embedding techniques map words into numerical vector spaces through an embedding matrix. Through the mapping, richer numerical representations of text input are created, enabling the deep multi-instance models to rely on these vector representations and improve performances in specific tasks. In our paper, the embedding takes a sequence as input, corresponding to a set of news titles. Assume that on a trading day $B_k, k = 1,2,\ldots,M$ with $\mathrm{n_k}$ news titles (instance), each news title $n_k^i$ contains $T_i$ words. $\{w_k^i\}^t, t \in [1, T_i]$ stands for the $t$th words in the $i$th news item of day $k$. We first embed the individual words $\{w_k^i\}^t$ to vectors through word embedding matrix:

$$L_w \in R^{d\times|V|}$$

where $d$ is the dimension of word vector and $|V|$ is vocabulary size. Then the embedded vectors for word $\{e_k^i\}^t \in R^d$ is gained through

$$\{e_k^i\}^t = L_w \times \{w_k^i\}^t$$

The word vectors can be either randomly initialized or be pre-trained with embedding learning algorithms such as Glove and Word2Vec. Here, we adopt Glove [47] for better use of semantic and grammatical associations of words. In details, the Glove file that pre-trained 100-dimension word vectors on 6 billion tokens, 400K vocabulary, has covered most of the words in our news texts.

### 3.2.2. News(instance) encoding

Drawing inspiration from [34, 36], we exploit a Bidirectional-LSTM after word embedding to incorporate the contextual information from both directions for words. The recurrent structure in LSTM promotes the capture of context information. Compared with standard recurrent neural network (RNN), the gated mechanism in LSTM prevents the unbounded cell state and tackles the problem of vanishing/exploding gradient, which makes it more applicable in modeling semantics of long texts. Hence, we have the following computation of LSTM cells:

$$f_t = \sigma\left(W_f\left[\{h_k^i\}^{t-1}, \{e_k^i\}^t\right] + b_f\right)$$

$$i_t = \sigma\left(W_i\left[\{h_k^i\}^{t-1}, \{e_k^i\}^t\right] + b_i\right)$$

$$\widetilde{C_t} = \tanh\left(W_C\left[\{h_k^i\}^{t-1}, \{e_k^i\}^t\right] + b_C\right)$$

$$C_t = f_t \otimes C_{t-1} + i_t \otimes \widetilde{C_t}$$

$$o_t = \sigma\left(W_o\left[\{h_k^i\}^{t-1}, \{e_k^i\}^t\right] + b_o\right)$$

$$\{h_k^i\}^t = o_t \otimes \tanh(C_t)$$

In LSTM, there are three gates, i.e. input gate $i_t$, forget gate $f_t$, and output gate $o_t$. For current input $\{e_k^i\}^t$ at time $t$ and previous hidden state $\{h_k^i\}^{t-1}$ at time $t-1$, the calculation in forget gate $f_t$ indicates the ability to forget old information. This gate decides what information should be forgotten or kept. Input gate $i_t$ is derived from input data $\{e_k^i\}^t$ and previous hidden node $\{h_k^i\}^{t-1}$ through a neural network layer. $\widetilde{C_t}$ represents the cell state update value. Through the forget gate and the input gate, the cell state $C_t$ is gained, with information of $C_{t-1}$ and $\widetilde{C_t}$. The output gate $o_t$ decides what the next hidden state $\{h_k^i\}^t$ should be. $\{h_k^i\}^t$ is obtained from the output gate $o_t$ and cell state $C_t$, where $o_t$ is calculated in the same way as $f_t$ and $i_t$. σ represents the sigmoid activation function.

The bidirectional LSTM contains the past and future context of the word. Through two hidden states $\overrightarrow{LSTM}$, $\overleftarrow{LSTM}$, information can be preserved, at any point in time, from both past and future. The forward $\overrightarrow{LSTM}$ makes news be read from the first word to the last word, and the backward $\overleftarrow{LSTM}$ allows information in news to flow from $\{w_k^i\}^{T_i}$ to $\{w_k^i\}^1$. Therefore,

$$\overrightarrow{\{h_k^i\}^t} = \overrightarrow{LSTM}\{e_k^i\}^t, \qquad t \in [1, T_i]$$

$$\overleftarrow{\{h_k^i\}^t} = \overleftarrow{LSTM}\{e_k^i\}^t, \qquad t \in [1, T_i]$$

We concatenated two hidden vectors $\overrightarrow{\{h_k^i\}^t}$ and $\overleftarrow{\{h_k^i\}^t}$ into

$$\{h_k^i\}^t = \left[\overrightarrow{\{h_k^i\}^t}, \overleftarrow{\{h_k^i\}^t}\right] \in R^{2\times u},$$

which represents $i$th news title in the $k$th day(bag). $n_k$ refers to the total amount of news items on $k$th day, and $u$ is hidden units of LSTM.

Words within the news are not equally informative to investors. Investors usually pay more attention to keywords whenever they see a news story. Hence, we introduce an attention mechanism on top of the Bi-LSTM layer, so that it can reward the words offering critical information in our news(instance) representation. In details:

$$\{u_k^i\}^t = \tanh\left(W_w\{h_k^i\}^t + b_w\right)$$

$$\{\alpha_k^i\}^t = \frac{\exp\left(\{u_k^i\}^t\right)}{\sum_t \exp\left(\{u_k^i\}^t\right)}$$

$$n_{ik} = \sum_t \{\alpha_k^i\}^t \times \{h_k^i\}^t \in R^{2u}$$

We output the news(instance) vector as a weighted sum of the encoder hidden states. Then we compute the attention scores $\{\alpha_k^i\}^t$ and take softmax to get attention scores into a probability distribution. Finally, we take a weighted sum of values using attention distribution, obtaining the attention output as our news(instance) vector. The attention output mostly contains information from the hidden states that received high attention. Thus, the news(instance) vector is beneficial to aggregate the representation for informative words and better focus on keywords within the news.

### 3.2.3. News(Instance)-level classifiers

After we obtain the dense representations $n_{ik}$ for each piece of news (instance) in bag $k$, the news-level classifiers, albeit labels are unobserved in the training set, are constructed to make predictions at the news(instance) level and infer the probability of each unseen individual news driving the stock up or down. For the classifiers of news(instances), we feed the news vectors $n_{ik}$ into a one-layer MLP with sigmoid activation:

$$\widehat{p_k^i} = sigmoid(W_{news} n_{ik} + b_{news})$$

where $\widehat{p_k^i}$ represents a real-valued score, demonstrating the predicted probability that an instance, i.e., one piece of news belongs to a particular class label. $W_{news}$, $b_{news}$ are the parameters of new-level classifiers.

### 3.2.4. Bag-level vector representation and classifiers

In classical MIL problems, once setting up instance-level classifiers to get inferred instance labels in the bag, bag labels can be derived from gathering its individual instances labels through *The aggregation functions*. For commonly used aggregation functions, maximum operation, mean or weighted averaging have been chosen in many previous works of literature [22, 23, 25, 26]. In fact, the instances vectors don't need to be processed any further at the most of the time. However, for financial news text classification problems, additional steps for the bag feature extraction and bag-level representation are still necessary.

To make multi-instance learning more suitable in the classification problems of financial news text, we are going to take a novel approach to learn vector representations of days(bag). To be specific, after we deduce the news(instance) probabilities from news(instance)-level classifiers, we do not directly aggregate them into the stock trend probabilities of a day(bag) and get bag-level predictions. Instead, we encode the possibility of its composed news(instance) into the day(bag) vector representation, and then build the day(bag)-level classifiers on top of that. In this case, a class of transformations from instances to bags can be parameterized by neural networks, making MIL more flexible and being trained end-to-end. Hence, in the following steps we have vector-based representations for day (bag) $k$:

$$z_k = \frac{1}{n_k} \sum_{i=1}^{n_k} \widehat{p_k^i}\, n_{ik}$$

Bag representation $z_k, k = 1,2, \dots, M$ is based on the prediction probability of its component new(instance). Using these vector representations, we can get the predicted day-level probabilities through day-level classifiers. By comparing the predicted day-level probabilities against the actual day labels, we can compute a cost function, and the network is then trained to minimize the cost. Detailed design of our proposed multiple instance learning model is displayed in Figure 3.

# 4. EXPERIMENTS

## 4.1. Datasets

To conduct our experiments, the dataset we use is a set of financial news released by Ding et al. [32], between October 2006 and November 2013 in daily frequency. This dataset contains 106,521 news from Reuters and 447,145 news from Bloomberg. According to [32, 36], news titles alone are more predictive than adding news contents for trend forecasting tasks. Therefore, we extract the publication timestamps, the title for each piece of news from this dataset for our experiment. To catch the time period of the financial news, the historical stock price data for shares in Standard & Poor's 500 (S&P 500) index at the same period are also collected from Yahoo Finance to conduct our experiment on forecasting tasks.

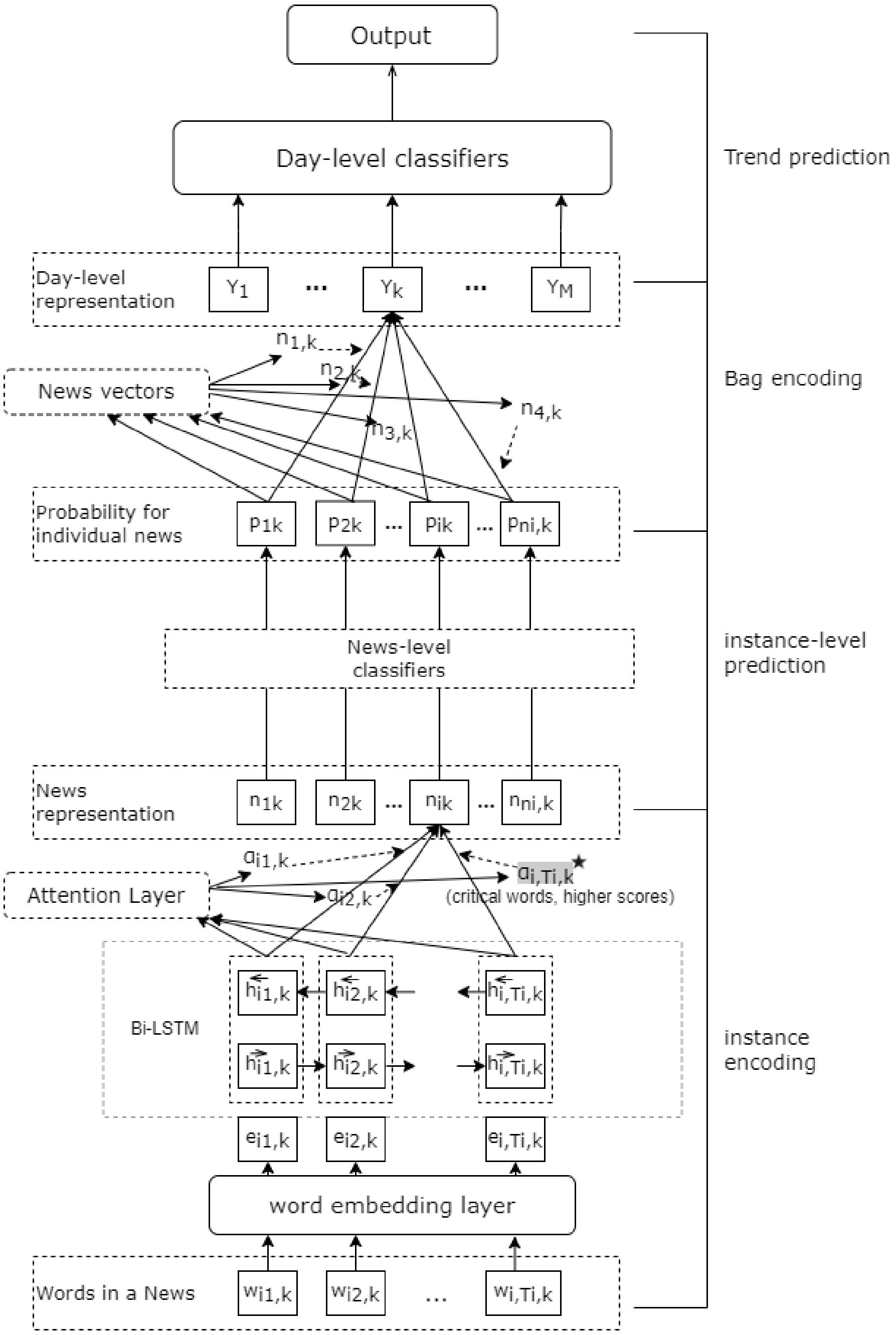

Figure 3. The overall architectures of our proposed Multiple Instance Learning Network

#### 4.1.1. Data pre-processing

In the following experiments, we conduct a series of basic text pre-processings such as lowercasing, tokenizing each news and removing the stop words and infrequent words (appearing less than 5 times). Subsequently, we filter out the news without any correlation to stocks, making sure that all related symbols and company names are included. After filtering, we obtained a total amount of 63403 news. Following text pre-processing, we split the dataset into training, validation and test set. Summary statistics of the training, validation and test set are in Table 2.

Table 2. Statistics Details of Datasets

| Datasets | Training | Validation | Test |
|---|---|---|---|
| Time period | 20/10/2006-27/06/2012 | 28/06/2012-13/03/2013 | 14/03/2013-20/11/2013 |
| News amounts | 38454 | 13237 | 11712 |
| Mean | 11.078795 | 11.127219 | 11.261783 |
| Std | 2.369729 | 1.834530 | 1.885941 |
| Min | 4.000000 | 4.000000 | 5.000000 |
| Max | 22.000000 | 19.000000 | 19.000000 |
| Distribution | 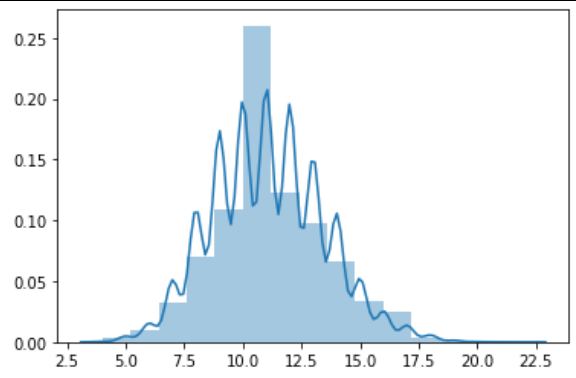 | 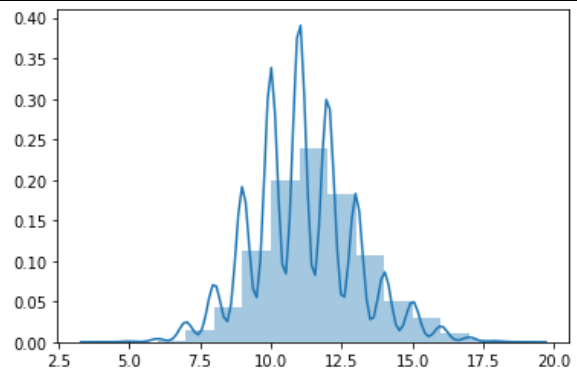 | 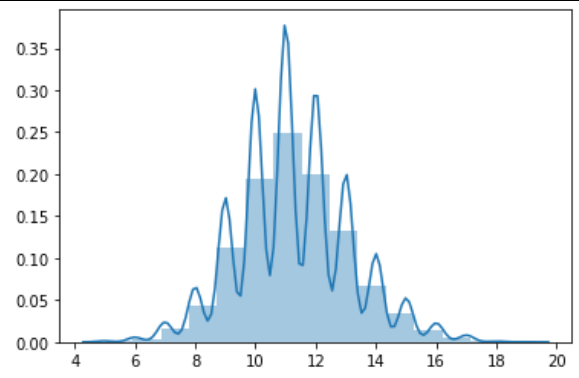 |

### 4.1.2. Experiment setup

To train our model, we use Adadelta algorithm as our optimization algorithm. Unlike the commonly used Stochastic Gradient Descent (SGD), the fixed global learning rate shared by all dimensions is less conducive to speeding up progress. The training progress can be slow when the gradient magnitude is small. Adadelta, as an optimization method using the adaptive learning rate, can converge faster and be used when training deeper and more complex networks. To further, we set the initial learning rate α as 0.1. Mini-batches of 32 is organized through the training process. In news (instance) representation, we adopt GloVe embeddings [47] as the pre-trained word embeddings, where the vector size of the word embedding is $|e| = 100$. The LSTM hidden vector dimensions for each direction were set to 50 and the attention vector dimensionality to 100. In the stage of the day-level (bag-level) prediction, the model convolves its input with news(instance) embeddings. For hidden layers within the news (instance) classifiers, the hidden units are set to 150. Additionally, to prevent overfitting, we adopt a dropout rate of 0.5 after both the instance level and the bag level.

## 4.2. Model Comparison

To evaluate our proposed model, in this section we set up a few baselines in contrast to our hybrid model. Our method is compared with preceding mainstream models and previous MIL models with differences in aggregation approaches for instances and bags (mean operation and encoding with instance-level results). For the sake of simplicity, the following notation identifies each model:

• BW-SVM: bag-of-words and support vector machines (SVMs) prediction model

• E-NN: structure events tuple input and NN prediction model, originally put forwards in Ding et al. [32]

• EB-NN: Event embedding input and NN prediction model in Ding et al. [33]

• S-NN: Following models in [30], taking the entire news corpus as the whole input. A mean operator is per formed on word embedding vectors within each piece of news as news vectors. On top of that, an averaging is added on news vectors in one day. A standard neural network is used as the prediction model.

• S-LSTM: The same vector representations as S-NN model, except to use LSTM as the prediction model instead of NN.

• Att-NN: Leverage a hierarchical attention network (HAN) analogous with the one in Yang [34]. Following Yang's HAN structure, we see each day as each document, news in one day as sentences in each document. A day representation is constructed by first building representations of news and then aggregating them into a day representation. Besides, a word-level attention layer and a new-level attention are added to differentiate keywords in each news and important news in one day. A standard neural network is used to make predictions.

• Att-LSTM: The same encoding and HAN's structure embeddings as Att-NN model except to use LSTM as the prediction model

• S-GICF: The multiple instance learning framework proposed in [25]. According to [25], an objective function

• ATT-GICF: Same multiple instance learning setup and GICF cost function as in S-GICF except to represent news(instance) by encoding news titles through Bi-LSTM and attention mechanism.

• MIL-s: Use averaging on word embedding vectors within each piece of news to represent news (instance) vectors. On top of that, the multiple instance learning model is exploited. News(instance)-classifiers are set to infer class possibilities of each individual news. Day(bag) representation is built using vectors representation and inferred stock trend probability of its component news(instance).

• MIL-rep: Our proposed model. Using Bi-LSTM and attention mechanism to encode each piece of news (instance). On top of that, news(instance)-classifiers are set to infer class possibilities of each individual news. Day(bag) representation is built using vectors representation and inferred stock trend probability of its component news (instance).

In order to compare the model performances, we use the classification accuracy as our evaluation and to prove if our approach can compete with the best state-of-art methods on our benchmark dataset. Table 3 shows the results compared with the above baseline models in the previous literature.

Table 3. Final results on the test dataset

| **Models** | **Test Accuracy (Max)** |
|---|---|
| BW-SVM | 56.38% |
| E-NN | 58.83% |
| EB-NN | 62.84% |
| S-NN | 57.92% |
| S-LSTM | 60.79% |
| ATT-NN | 57.59% |
| ATT-LSTM | 61.93% |
| S-GICF | 58.52% |
| ATT-GICF | 59.09% |
| MIL-s | 60.23% |
| MIL-rep | **63.06%** |

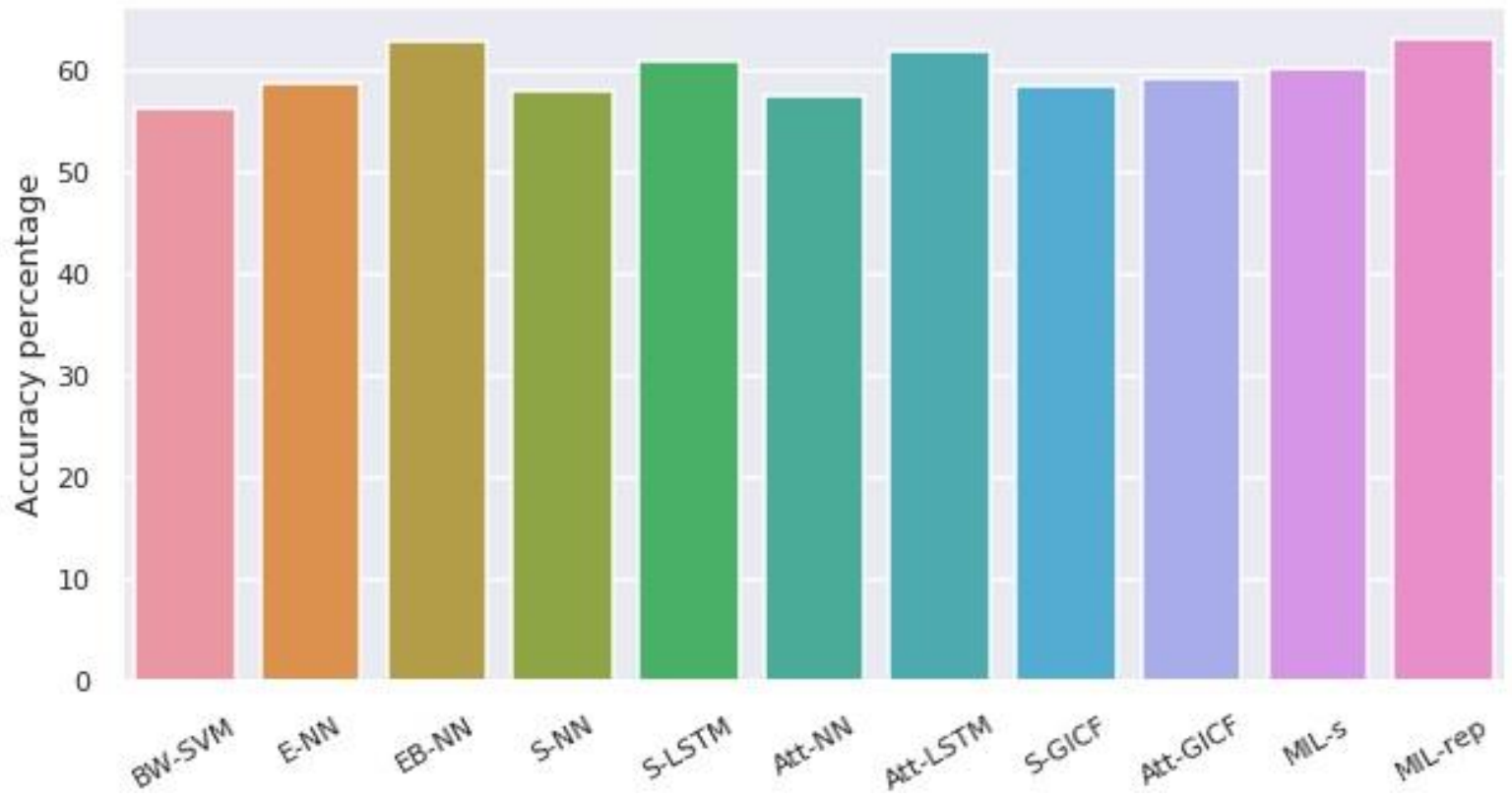

Figure 4. Performances in different models

## 4.3. Discussion

Through all the experiments above, our proposed framework, MIL-rep, can achieve a predominant accuracy of 63.06% (see in Table 3 and Figure 4), outperforming all the baseline models. Although event embedding (EB) in EB-NN is competitive, our proposed multiple instance learning framework is still powerful with event embedding (EB) in the extraction of financial news for stock trend forecasting. By making the comparison, we conclude the following discussions with respect to the following aspects:

**Discussion on news representations:** For the expression of news, in the non-MIL methods, we use simple averaging embedding (S) and encoding with Bi-LSTM, attention mechanism (ATT) to get the vector representation of news. In the MIL methods (S-GICF, ATT-GICF, MIL-s, MIL-rep), we take daily news as an instance. The representation of news is the instance representation. For MIL models, the results on the comparison between the models S-GICF vs ATT-GICF, MIL-s vs MIL-rep lead to the conclusion that better performances can be derived from the news(instance) representation of Bi-LSTM and attention (ATT) encoding, than the simple average embedding. In terms of the non-MIL method, in comparison with models S-LSTM vs ATT-LSTM, it can be inferred that ATT encoding is slightly better than simple averaging as input of the model, obtaining higher accuracy. Although something special in ATT-NN for case S-NN vs ATT-NN, we will discuss it later. It is not hard to see that the ATT representation for news is conducive to predicting the stock trend using financial news, especially in MIL models. This can be explained by the fact that the input to the model is organized in sequential contexts through Bi-LSTM. Keywords that show important trend signals in the news title are greatly extracted by the attention mechanism.

**Discussion on predictive models:** We mainly take LSTM as our primary predictive model in the experiments. In comparisons for non-MIL models, i.e., S-NN vs S-LSTM; ATT-NN vs ATT-LSTM, different predictive neural networks are constructed under the same news representation input. From Table 3, we can see that S-LSTM outperforms S-NN, and ATT-LSTM is better than ATT-NN. The structure in LSTM is effective at capturing long-term temporal features of the input sequences, which helps to enhance performances for the index prediction task.

**Discussion on effects of multi-instance learning:** The variable number of financial news (instances) in one trading day (bag) and the lack of news (i.e., instance-level) labels hinder us from inferring the stock trend labels of new days(bags). That's the reason why we move to multiple instances learning to address news(instance) level predictions. There are two kinds of multiple instance frameworks we adopt in the proposed tasks. One is MIL composition using GICF cost function from [25], another is the MIL construction we put forward in this paper.

In S-GICF and ATT-GICF models, we use the MIL methods with GICF cost function. Based on the intuition that news(instances) pairs with higher similarity in one day will be more likely to assign the same labels, news(instance) similarity is exploited to read the combination possibility and assign news(instance) labels in a day(bag). After that, a simple averaging aggregation function is chosen to gather instance labels as the predicted day(bag) labels. From the possible assignment of its individual instance labels, GICF model is quite suitable to solve the news uncertainty, as previously mentioned, in both random appearing and unknown combination aspects. However, this approach is instance-label-oriented, without further vector operations at bag levels. The generalization capability of the averaging aggregation function is relatively poor. There is a risk the model is not adequately trained. Labels gathering may omit some potential information through the averaging aggregation, failing to identify complex patterns in the proposed task. Take simple averaging, ATT-encoding approach as instance representation and separately calculate pair similarity, the results in S-GICF and ATT-GICF are modest, with less than 60\% accuracy. Compared with non-MIL methods, the MIL composition with GICF in performances of S-GICF, ATT-GICF are also not as good as the LSTM structures in S-LSTM, ATT-LSTM.

In our proposed MIL construction, we determine to use a joint representation of a bag instead of gathering labels of instances. Predicted by instance-level classifiers, the probability of the class to which each instance(news) belongs is used to encode into the bag vectors. In this way, our proposed model can not only establish bag-level classifiers, allowing full, end-to-end training, but also include instance features of news uncertainty. According to the results of MIL-s and MIL-rep, the fitting effects of financial news are greatly enhanced by this embedding manner. The learning and generalization capability of our MIL models (MIL-s and MIL-rep) are improved, better than not only S-GICF and ATT-GICF models but also other non-MIL baselines.

**Discussion on hierarchical attention in ATT-NN:** In ATT-NN, two levels of attention mechanisms are used to identify keywords in each news text and significant news in a trading day. However, the performances of ATT-NN are apparently worse than the others. ATT-NN model has hierarchical structures, embed vectors in common with ours. In order to distinguish ATT-NN model from ours, we use the predicted class probability of news(instance) to encode the day representations. To get the predicted class probability of each news text, we set up the news-level classifiers. In contrast to our approach, Att-NN continues to encode the day vectors through GRU and attention mechanism on top of news vectors, without establishing news-level classifiers. Furthermore, in compliance with HAN's structure, we automatically assume that there exists a strong correlation between each news item within one day. All news items in a day are highly context-dependent, i.e., the same news may be differentially important in a different context. However, the assumption can be too strong for news, which leads to poor performances and inferior outcomes in the training of this model. Since in real life, the news is relatively independent of each other. There is no semantic, context relationship between news pairs on the same day as there is between sentences in one document. Therefore, HAN's structure is more suitable in document modeling than financial news analysis. For our model, we reduce this sensitivity and treat news every day as relatively independent variables, with their effects being related and synergistic to the day's prediction performances. The experiment results show that our approach has the merits of high learning efficiency, high classification accuracy, and high generalization capability.

## 5. CONCLUSION AND FUTURE WORK

In this paper, we point out the challenges in dealing with the financial news in the stock trend prediction, particularly, the uncertainty of financial news in terms of random daily occurrences and unknown individual labels composition. To address these issues, we propose to adopt the principle of multi-instance learning, and solve the problem of news-oriented stock trend forecasting from the perspective of multiple instance learning for the first time. With this end in view, an adaptive, end-to-end MIL framework is developed in this paper to achieve better performances. Experimental results on S&P 500 index demonstrate that our proposed model is powerful and effectively increases performance. Nowadays, multi-instance learning is receiving moderate popularity for its applicability in learning problems with label ambiguity. In a variety of application scenarios, some supervised learning methods have also been included or extended to the MIL setting. Nevertheless, there is still much need to explore in the field of financial forecasting for multi-instance learning, which could be a direction for our future research. In addition, in this paper, we only focus on the impact of news on stock trends. Taking more online data resources from social media into account in the trend prediction model is also a potential research site in the future.